\documentclass[a4paper, conference]{IEEEtran}
\IEEEoverridecommandlockouts
\usepackage{cite}
\usepackage{amsmath,amssymb,amsfonts}
\usepackage[top=0.75in, bottom=1.00in, left=0.625in, right=0.625in]{geometry}
\usepackage{algorithmic}
\usepackage{graphicx}
\usepackage{textcomp}
\usepackage{xcolor}
\usepackage[acronym, nonumberlist]{glossaries}
\usepackage[hidelinks]{hyperref}
\usepackage[all]{hypcap}
\usepackage[english]{babel}
\usepackage{fancyhdr}

\begin{document}


\title{\vspace*{-0.375in}Development of a Low-Cost Prosthetic Hand \\Using Electromyography and Machine Learning
}


\author{
\IEEEauthorblockN{Mosab Diab\IEEEauthorrefmark{1}\IEEEauthorrefmark{2}, Ashraf Mohammed\IEEEauthorrefmark{1}\IEEEauthorrefmark{2}, Yinlai Jiang\IEEEauthorrefmark{2}, and Sharief Babikir\IEEEauthorrefmark{1}\IEEEauthorrefmark{3}}
\IEEEauthorblockA{\IEEEauthorrefmark{1}\textit{Department of Electrical and Electronic Engineering, Faculty of Engineering}, \textit{The University of Khartoum}, Sudan}
\IEEEauthorblockA{\IEEEauthorrefmark{2}\textit{Center for Neuroscience and Biomedical Engineering}, \textit{The University of Electro-Communications}, Chofu, Tokyo, Japan}
\IEEEauthorblockA{\IEEEauthorrefmark{3}\textit{Department of Electrical Engineering}, \textit{The University of Tennessee at Chattanooga}, The United States of America}
}

\maketitle



\thispagestyle{fancy}

\begin{abstract}
Electromyography (EMG) is a measure of muscular electrical activity and is used in many clinical/biomedical disciplines and modern human computer interaction. Myo-electric prosthetics analyze and classify the electrical signals recorded from the residual limb. The classified output is then used to control the position of motors in a robotic hand and a movement is produced. The aim of this project is to develop a low-cost and effective myo-electric prosthetic hand that would meet the needs of amputees in developing countries. The proposed prosthetic hand should be able to accurately classify five different patterns (gestures) using EMG recordings from three muscles and control a robotic hand accordingly. The robotic hand is composed of two servo motors allowing for two degrees of freedom. After establishing an efficient signal acquisition and amplification system, EMG signals were thoroughly analyzed in the frequency and time domains. Features were extracted from both domains and a shallow neural network was trained on the two sets of data. Results yielded an average classification accuracy of 94.8\% and 93.2\% for the time and frequency domains respectively. Furthermore, results showed a faster computation and response for the time domain analysis; hence, it was adopted for the classification system. A wrist rotation mechanism was designed and tested to add significant functionality to the prosthetic. The mechanism is controlled by two of the five gestures, one for each direction. Which added a third degree of freedom to the overall design. Finally, a tactile sensory feedback system which uses force sensors and vibration motors was developed to enable sensation of the force inflicted on the hand for the user.
\end{abstract}


\section{Introduction}
\begin{figure}
         \centering
         \includegraphics[width=0.485\textwidth]{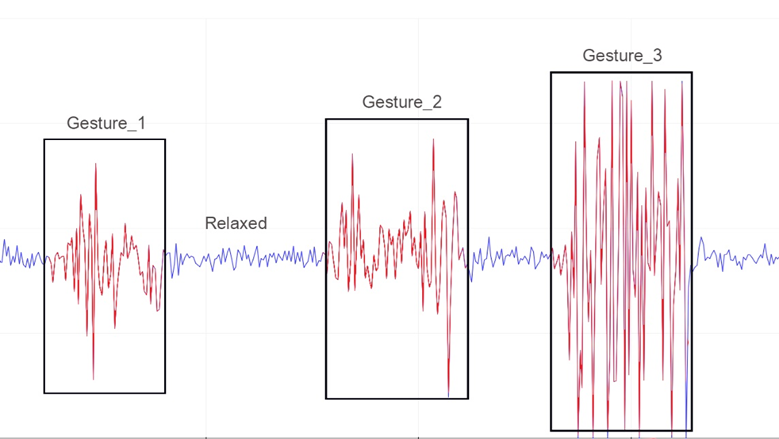}
        \caption{EMG time domain signal of relaxed versus contracted muscle}
        \label{fig:EMG_Signal}
\end{figure}

The World Health Organization estimates that there are 40 million amputees in developing countries. Shockingly, only 5\% have access to any type of prosthetics due to their expensiveness and unavailability \cite{Limbs2018}. Moreover, the prosthetics available in these developing countries are of low quality, uncomfortable, and are inefficient. Although there have been attempts to develop low-cost prosthetics in developing countries, amputees’ demands are yet to be met \cite{Marino2015}. This project aims to develop a low-cost prosthetic hand that is controlled by the muscular electrical signals recorded from the residual limb. The prosthetic should be able to perform key grips in addition to wrist mobility at a high accuracy and have a fast response. Furthermore, the prosthetic should be able to send feedback signals to the residual limbs indicating how much pressure is applied to the fingers. ``Electromyography (EMG) is an experimental technique concerned with the development, recording and analysis of myoelectric signals. Myoelectric signals are formed by physiological variations in the state of muscle fiber membranes." \cite{Konrad2005}. Whenever a hand movement is intended the brain sends an electrical signal through the motor unit to the muscle fibers. These electrical signals differ from one movement to the other; this allows for hand movement classification through pattern recognition of Surface EMG (SEMG) signals. Figure \ref{fig:EMG_Signal} shows an EMG signal produced when relaxing the muscle versus performing three different hand gestures.

Any implant or prosthetic replacing a function or functions of an organ or a group of organs should be biologically and sensorily integrated with the human body in order to increase their acceptance with their user. If this replacement is for a human hand which is an important interface between humans and their environment, the acceptance issue and developing sensory-motor embodiment will be more challenging. Despite progress in prosthetics’ technologies, a limited number of hand amputees wear a prosthetic device. In this paper, we propose a prosthetic hand that is easily accessible due to its simplicity, yet efficient due to its accuracy.
We study different methods of processing EMG signals for myoelectric prosthetic hands. We also develop a fully functional myoelectric prosthetic hand using these methods. In the development, we focus on reducing the complexity of the system in both hardware and software whilst maintaining high-level functionality.


\section{Myoelectric Signals}

\subsection{Signal Acquisition}

Biopotential\footnote{Biopotential electrodes are transducers that convert the muscles’ ionic current into the traditional current flowing in the electrode.} \cite{Tbeh2006} surface electrodes are used to conduct the small currents from the muscle to the pre-processing units and could be grouped into wet and dry electrodes:\\
Wet Electrodes: Ag/AgCl electrodes with wet conductive gel acting as an electrolyte are traditionally used for EMG signal acquisition by adhering to the body. Their characteristics are well described and yield accurate results \cite{Chi2010}. The downside of using Ag/AgCl is that their impedance increases with use and must be disposed of after a few uses.\\
Dry Electrodes: Can be used multiple times yielding same accuracy and are used in long term applications such as prosthetics. Conductive materials like certain metals and conductive fabric can be used as dry electrodes. Moreover, different resin mixes can be used to produce electrodes with a multitude of characteristics. Unlike Ag/AgCl electrodes, they don’t adhere to the body and positioning should be mechanically addressed.

\subsection{Signal Pre-Processing}

Raw EMG signals range between +/- 5000 microvolts and their frequency contents range between 6 and 500 Hz \cite{Konrad2005}. And so, two techniques must be applied, which are: amplification using instrumentation amplifiers and noise cancellation using special circuitry and filters.

\subsection{EMG Signals Analog to Digital Conversion (ADC)}

The sampling rate and the ADC resolution and quantization are the two factors affecting EMG ADC. Applying the Nyquist sampling theorem, Sampling with lower frequencies will result in signal aliasing (information loss). Since EMG signals’ highest frequency component naturally is around 500 Hz, a sampling rate of at least 1000 Hz must be applied. Digitized signals are an approximation of the waveform analog signals; the ADC resolution is how many intensity levels the signals get approximated to (quantization). The higher the resolution the more accurate the conversion is. However, ADC resolution is limited by hardware, and normally, ADC chips with higher resolutions are more costly; Hence, a trade-off between cost and accuracy must be made.

\subsection{Signal Analysis and Feature Extraction}

The choice of feature space\footnote{Features are calculated using a moving window; taking a certain number of samples for a few milliseconds and performing the calculations on them.} is dictated by how well it can cluster and categorize input data. In addition, the chosen feature space should meet the hardware limitations of the processing unit chosen for the certain application; hence, should be computationally low cost. Finally, the feature space should be able to maintain separability of input clusters under noisy environments \cite{Boostani2003}. Feature spaces can be mainly categorized into Time Domain Features (TDF) and Frequency Domain Features (FDF). Because EMG signals are time varying with random positive and negative values, it intrinsically holds time domain information. TDFs are widely used in the field of research and engineering because of the high-quality classification they provide and their low computational complexity \cite{Gulten2017}. Furthermore, because no transformation is needed, time domain features maintain the integrity of the data. The following features are commonly used in gesture classification \cite{Altin2016}:

\begin{itemize}
    \item Mean: A common statistical measure taking the average value of the window.
    \begin{equation}\label{eq:mean}
        \mu=\frac{1}{N}\sum_{n=1}^{N}x_{n},
    \end{equation}
\end{itemize}
Where $x_{n}$ is the data point and $N$ is the total number of data points.

\begin{itemize}
    \item Variance: Also, a common statistical measure performed on the moving window.
    \begin{equation}\label{eq:variance}
        var=\frac{1}{N-1}\sum_{n=1}^{N}(x_{n}-\mu)^{2}
    \end{equation}
    \item Standard Deviation:
    \begin{equation}\label{eq:std}
        \sigma=\sqrt{var}
    \end{equation}
    \item Kurtosis: Kurtosis is a measure of the sharpness of the probability distribution peak.
    \begin{equation}\label{eq:kurtosis}
        kurt=\frac{\frac{1}{N}\sum_{n=1}^{N}(x_{n}-\mu)^{4}}{\sigma^{4}}
    \end{equation}
    \item Waveform Length: Waveform length is a measure of complexity of the EMG signal. It is defined as the cumulative length of the EMG waveform over the time segment \cite{Altin2016}.
    \begin{equation}\label{eq:WL}
        WL=\sum_{n=1}^{N-1}\lvert x_{n+1}-x_{n}\rvert
    \end{equation}
\end{itemize}

Frequency domain analysis means finding the sinusoidal harmonics which EMG signals consist of. The power distribution of the signal across the frequency spectrum is found and corresponding amplitudes can then be calculated and used as features.  Mean and median frequency can also be used as features for further analysis. Fast Fourier Transform (FFT) is the easy and common way to convert the signals to the frequency domain \cite{Kanwade2016}.

\subsection{Classification}

After analyzing and extracting the features from the EMG signals, they can now be used to classify movements that correspond to feature patterns (clusters). The most common classification algorithms are K-Nearest Neighbor Algorithm (KNN), Linear Discriminant Analysis (LDA), Artificial Neural Networks (ANN) and Support Vector Machines (SVM) \cite{Altin2016}.

\subsection{Feedforward Artificial Neural Networks (FFANN)}

A computational architecture consisting of an input layer, hidden layer/s, and an output layer. Each layer is defined by weights and bias matrices connected to it, and a transfer function that determines the output of that layer. The objective is to train this architecture to be able to successfully classify input data. This is done by updating the weights and biases to reduce the cost function. FFANN are very flexible; by increasing/decreasing the number of hidden layers and changing the transfer functions, results change drastically. In addition, they are considered computationally low cost when classifying the data \cite{Hagan2014}.

\section{Methodology}

\subsection{EMG Signal Processing}

\subsubsection{Amplification}

The AD620 IC instrumentation amplifier was used with the differential gain set at 100, and the AD8618 IC containing four operational amplifiers was used for amplification and as a notch filter to filter out the 50Hz power line noise. The CMRR of the amplifier is 64.8dB.

\subsubsection{Gesture Classification System}

The EMG signals were recorded from three muscles in the forearm (input channels) using special purpose electrodes and the amplifiers mentioned earlier.

\begin{figure}
         \centering
         \includegraphics[width=0.483\textwidth]{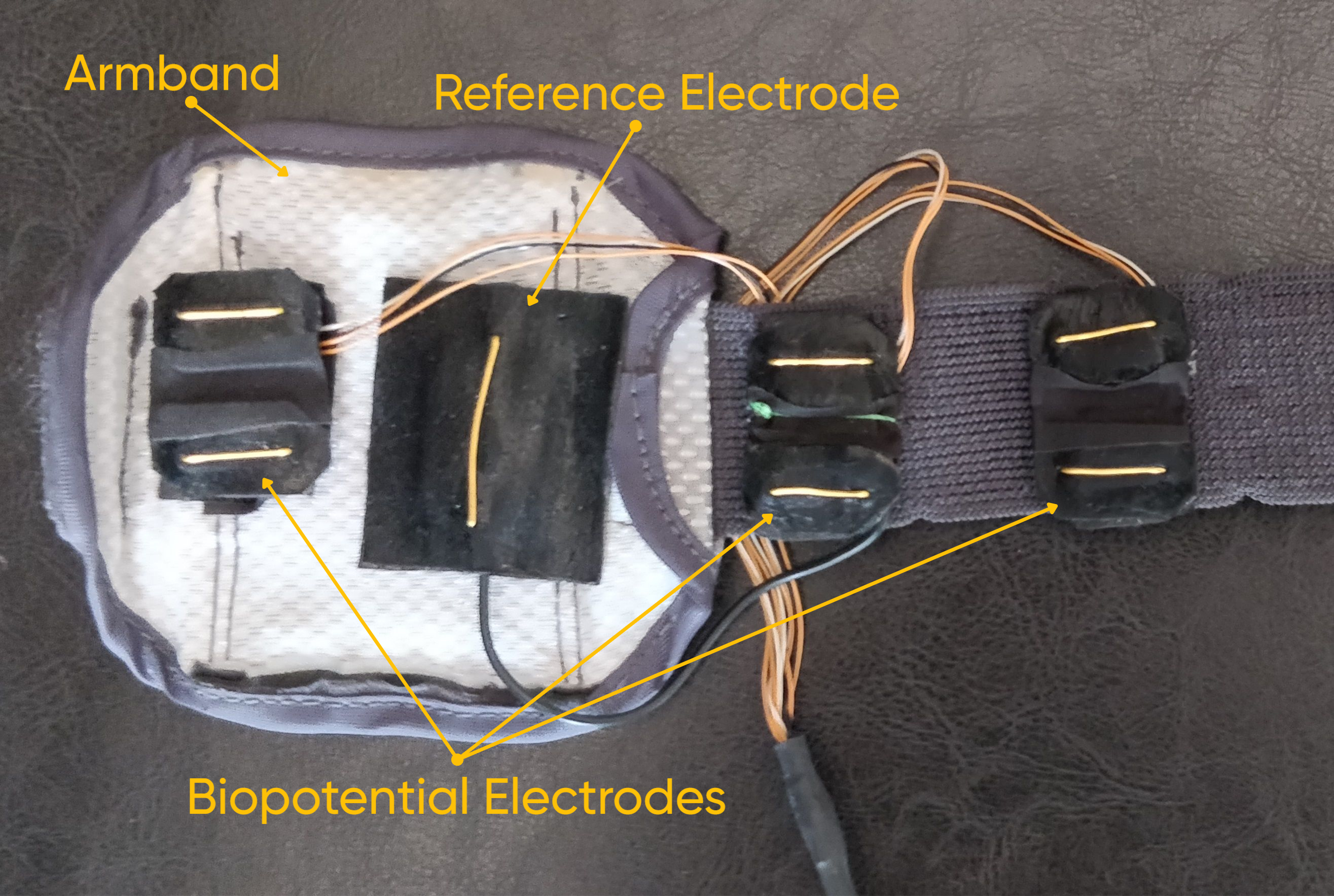}
        \caption{EMG electrodes}
        \label{fig:EMG_Electrodes}
\end{figure}

\begin{itemize}
    \item Choice of Muscles: The three channels were chosen arbitrarily from the upper forearm following the general guidelines for EMG recording as mentioned in \cite{Konrad2005}.
    \item Gestures: The chosen gestures are, open, grasp (Power Sphere as shown in \cite{Feix2016}), extension, flexion, and relaxed.
    \item Gesture Set and Recording Protocol: Five gesture classes were recorded: hand open, hand close (grasp), clockwise wrist rotation (extension), counter-clockwise wrist rotation (flexion), and rest (relaxed). Each gesture was performed for 10 seconds per trial. Trials captured both transient and steady-state muscle activity: each began from a relaxed baseline, with participants either transitioning from rest into a gesture (rest to open, rest to close) or holding a gesture in isolation. The dataset therefore contains onset transients, sustained contractions and rest segments rather than steady-state activity alone. This was intentional, as a classifier trained only on sustained contractions degrades across the transition periods that dominate real-time use, when the user is moving between gestures rather than holding one. Windows spanning a gesture onset were labeled.
    \item Electrodes: Special electrodes consisting of two carbon-silicon layers, a conductive fabric layer, and a gold coated copper wire were used to read the EMG signals. The electrodes are shown in Figure \ref{fig:EMG_Electrodes}.
    \item Controller: The microcontroller used for the gesture classification and control of the prosthetic hand was an Arduino Mega 2560 with 256 KB of flash memory.
\end{itemize}

\paragraph{Frequency Domain}

\begin{figure}
         \centering
         \includegraphics[width=0.485\textwidth]{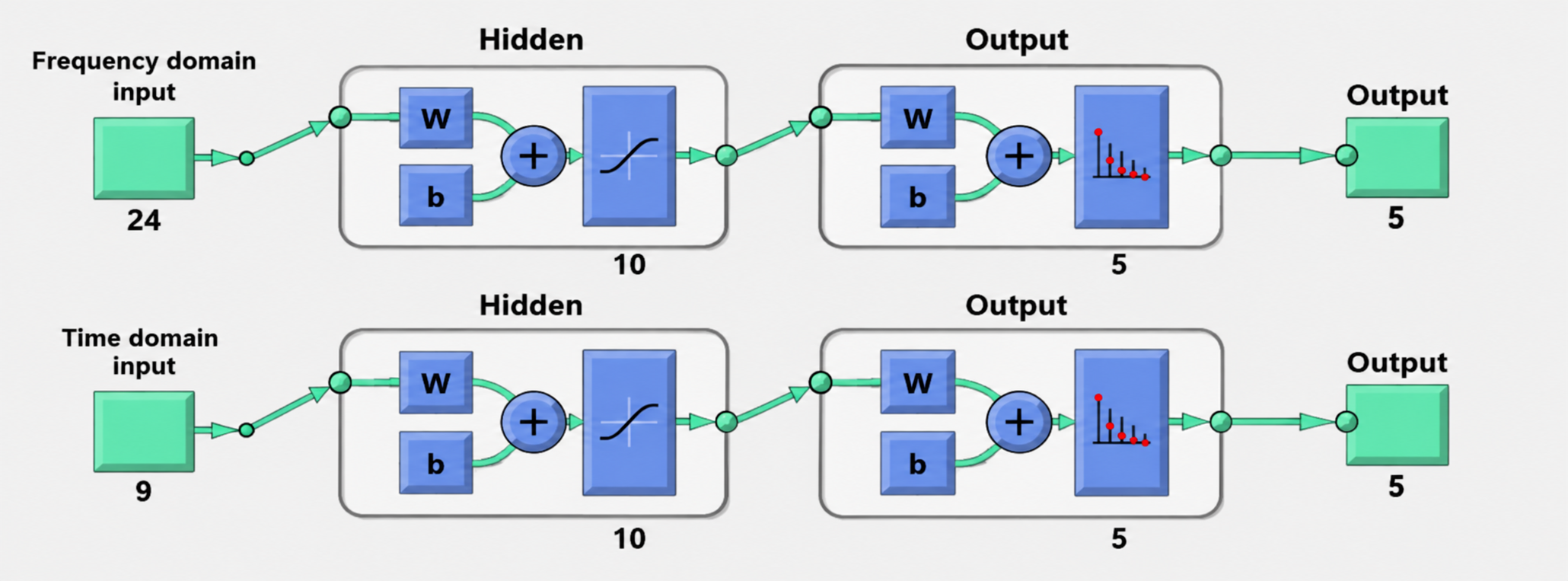}
        \caption{Frequency (top) and time (bottom) domain neural network structure}
        \label{fig:Combined_ANNs}
\end{figure}

\begin{itemize}
    \item Feature Extraction: To ensure that no aliasing will occur to the data, the sampling rate was set to 2000 samples/second. This will generate a frequency spectrum from 0 to 1000Hz. The Fast Fourier Transform (FFT) has been used to convert the signal to the frequency domain using 128 samples per window. This will divide the frequency spectrum into 64 bins, each bin representing (1000/64) 15.6Hz. It should be noted that the number of samples/window should be a power of 2 to keep the FFT’s computation at minimum complexity. Moreover, the accuracy of the conversion is proportional to the number of samples/window. However, Arduino memory limitations were considered, and 128 samples gave the best results while meeting hardware limitations. To reduce computational complexity and maintain accuracy, eight frequency bins for each channel have been chosen to represent the neural network’s input features. The 8 bins corresponded to the highest power values in the frequency distribution.
    
    \item Neural Network Classification: A feedforward artificial neural network consisting of a 24-neuron\footnote{Eight features for each of the three channels.} input layer, 10-neuron hidden layer with tansig transfer function, and a 5-neuron output layer with softmax transfer function was used to classify the input data. The network was constructed and trained using MATLAB. The output of the trained network gave the probability of each class. Intuitively, the highest probability corresponds to the classified gesture. Figure \ref{fig:Combined_ANNs} shows the structure of the neural network.
\end{itemize}

\paragraph{Time Domain}

\begin{itemize}
    \item Feature Extraction: A sampling rate of 2000 samples/second and a moving window of 128 samples was also adopted in the time domain analysis. The following properties were then calculated:
    \begin{itemize}
        \item Variance: A measure of how far the samples are spread out from the mean as defined in \eqref{eq:variance}.
        \item Mean Absolute Deviation (MAD): The average of the absolute deviations of data points from their mean.
        \begin{equation}\label{eq:MAD}
            MAD=\frac{1}{N}\sum_{n=1}^{N}\lvert x_{n}-\mu \rvert
        \end{equation}
        \item Waveform Length as defined in \eqref{eq:WL}.
    \end{itemize}
\end{itemize}

\begin{figure}
         \centering
         \includegraphics[width=0.485\textwidth]{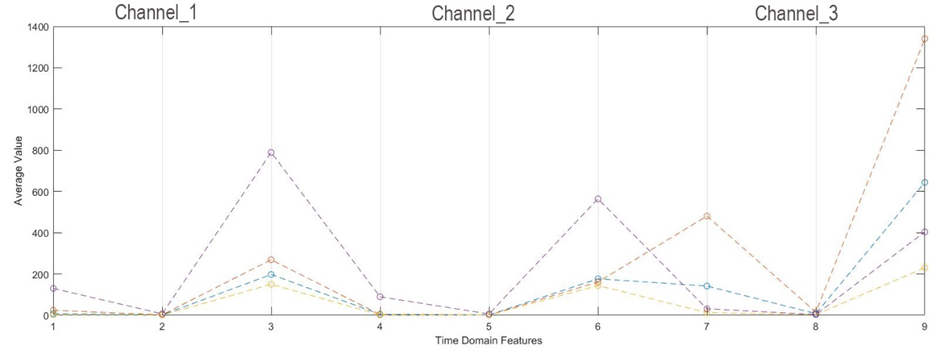}
        \caption{Time domain features}
        \label{fig:Time_Features}
\end{figure}

Figure \ref{fig:Time_Features} shows a graphical representation of the averaged time domain features’ values for four of the five hand gestures. It also shows how the features vary for each hand gesture.

\begin{itemize}
    \item Neural Network Classification: The same neural network structure was used with nine neurons\footnote{Three features for each of the three channels.} as input layer and tansig and softmax as transfer functions. Figure \ref{fig:Combined_ANNs} shows the time domain neural network structure.
\end{itemize}

\subsection{Dataset Composition}

Both training sets were class-balanced across the five gestures: 200 windows per gesture for the time-domain model (1000 windows total) and 145 per gesture for the subset used in the frequency-domain analysis (725 windows total). We held-out evaluations using independent recordings, again class-balanced, comprising 130 windows per gesture for the time-domain model (650 total) and 100 per gesture for the frequency-domain model (500 total). Training and validation windows were separated at the trial/session level, such that no two windows originating from the same continuous recording appeared in both sets.

\subsection{Prosthetic Hand Mechanism}
The prosthetic hand was 3D printed using polylactic acid and contains two servo motors. One of the motors is connected to the thumb and the other to the four remaining fingers allowing for two degrees of freedom.

\begin{figure}[!t]
         \centering
         \includegraphics[width=0.2\textwidth]{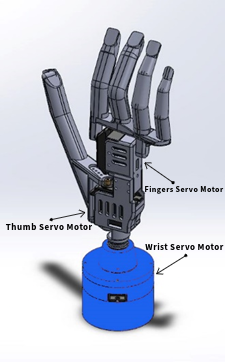}
        \caption{Mechanical design of the prosthetic hand}
        \label{fig:Hand_Design}
\end{figure}

To increase the effectiveness of the developed prosthetic hand, a mechanism that allows for user-controlled wrist rotation was designed using servo motor steel ball bearings, and custom-made 3D printed parts. This added a third degree of freedom to the overall design. Figure \ref{fig:Hand_Design} shows the mechanical design for the prosthetic hand with the wrist rotation mechanism highlighted in blue.

The three servo motors rotate depending on the gesture provided by the user to achieve the required motion of the prosthetic hand. The motors were controlled by hardware interrupts initiated by the classification of a certain gesture and the corresponding action is then performed by the motors. Four different movements can be performed by the robotic hand which are: open, grasp, clockwise, and counter-clockwise rotation. This rotation happens at a 180-degree angle. The hand can stop during any movement when the corresponding gesture is classified.

\subsection{Tactile Sensory Feedback System}

\subsubsection{Force Sensing Resistors (FSRs)}
\begin{figure}
         \centering
         \includegraphics[width=0.485\textwidth]{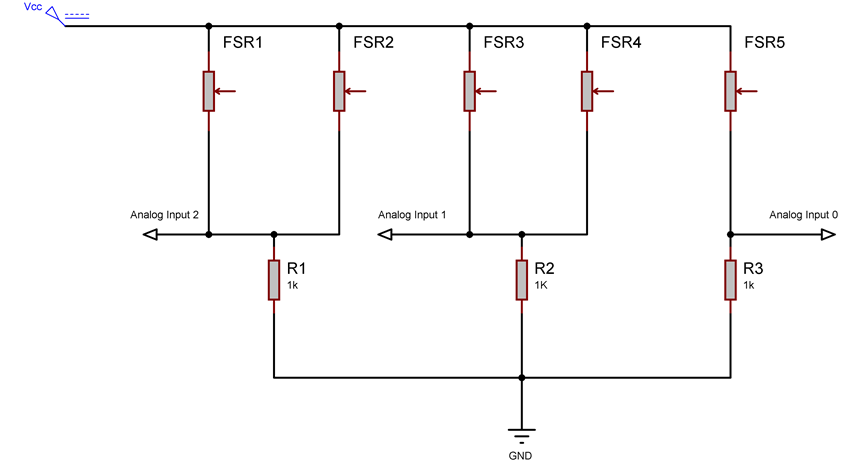}
        \caption{Force sensing circuit design}
        \label{fig:Sensory_Feedback_System}
\end{figure}

Force sensing resistors are used on each of the fingertips of the prosthetic hand to sense the force on them. The higher the force on the FSR the lower its resistance is. If the force on the FSRs is higher than the maximum threshold, then the vibration motors will remain operating at maximum power. The hand was divided into three regions with each region corresponding to one of the three vibration motors, the circuit design for the force sensing is shown in Figure \ref{fig:Sensory_Feedback_System}.

\subsubsection{Vibration Motors}
Coin vibration motors were used to provide feedback from the FSRs. Three motors were used with each motor operated between 0-2.5V with 2.5V being maximum vibration.

\begin{figure}[!t]
         \centering
         \includegraphics[width=0.483\textwidth]{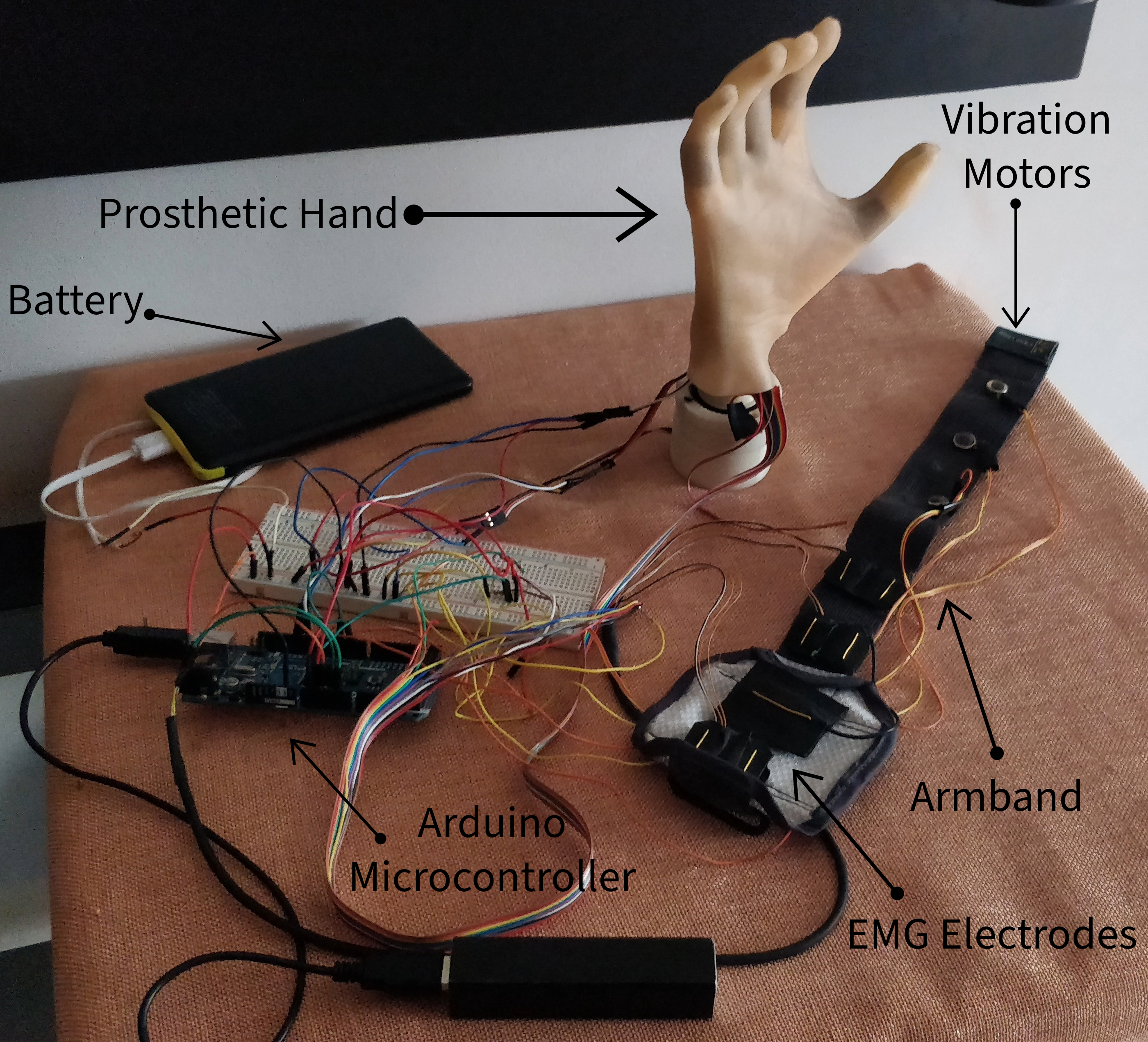}
        \caption{Full prosthetic hand system}
        \label{fig:Final_System}
\end{figure}

Figure \ref{fig:Final_System} shows the finalized prosthetic hand system.

\section{Results}

\subsection{Test Group}

The test group consisted of five healthy and able-bodied participants due to availability and social context. The group contained three men and two women. All between the ages of 21 and 26. The participants were planned as base models before testing on real users.

\subsection{Gesture Classification}

The results were measured in the frequency and time domain analysis and the accuracy of the neural network was calculated as follows:

\begin{equation}\label{eq:accuracy}
        Accuracy=\frac{Number \, of \, Correct \, Classifications}{Total \, Number \, of \, Classifications}
    \end{equation}

\subsubsection{Frequency Domain Classification Accuracy}

\begin{figure}
         \centering
         \includegraphics[width=0.485\textwidth]{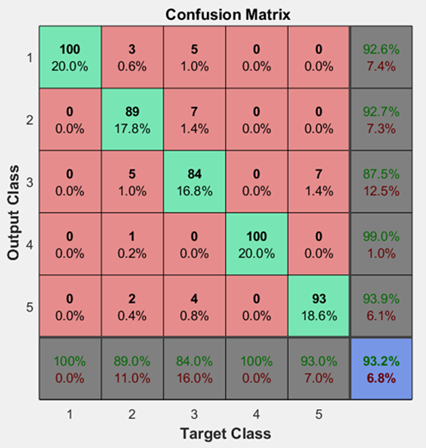}
        \caption{Frequency domain testing data confusion matrix}
        \label{fig:Frequency_Matrix}
\end{figure}

The network was trained on 725 data entries. The accuracy obtained was 98.6\% for the training data, 94.4\% on the validation data, and 100\% on the initial testing data. Further testing on another data set containing 500 data entries was conducted and the results are shown in Figure \ref{fig:Frequency_Matrix}. The accuracy obtained was 93.2\%.

\subsubsection{Time Domain Classification Accuracy}

\begin{figure}
         \centering
         \includegraphics[width=0.485\textwidth]{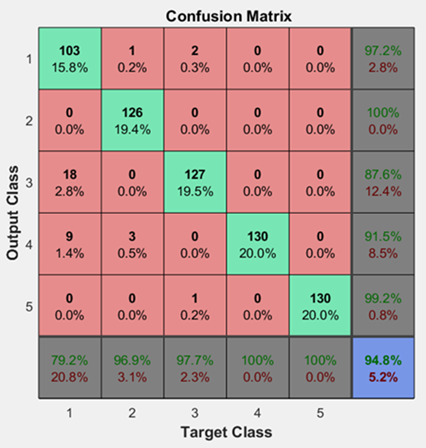}
        \caption{Time domain testing data confusion matrix}
        \label{fig:Time_Matrix}
\end{figure}

The network was trained on 1000 data entries and yielded accuracy of 99.7\% for the training data, 100\% on the validation data, and 100\% on the initial testing data. Further testing on another data set containing 650 data entries was conducted and the results are shown in Figure \ref{fig:Time_Matrix}. The accuracy obtained was 94.8\%.

\subsection{Comparison Between Frequency Domain and Time Domain Analysis}

Table \ref{table:comparison} shows the comparison of the two types of analyses. The comparison was made under the same sampling rate and window size. Both analyses were done on pre-stored training and testing datasets.
\begin{table}[!t]
\caption{Comparison between frequency domain and time domain analyses}
\centering
\begin{tabular}{|p{0.33\linewidth}||p{0.25\linewidth}|p{0.2\linewidth}|}
 \hline
 Parameter & Frequency Domain & Time Domain \\
 \hline\hline
 Average Accuracy & 93.2\% & 94.8\% \\ 
 \hline
 Algorithm Complexity & O(nlogn) & O(n) \\
 \hline
 \raggedright Feature Extraction Time/Window (ms) & 198 & 26 \\
 \hline
 Training Epochs & 70 & 99 \\
 \hline
 Training Time (Seconds) & 1 & 33 \\ 
 \hline
\end{tabular}
\label{table:comparison}
\end{table}

\section{Results Analysis and Comparisons}

The authors in \cite{Basak2021} experimented with Support Vector Machines (SVM) and ANNs for classifying EMG signals and used Principal Component Analysis (PCA) and Linear Discriminant Analysis (LDA) for analysis and dimensionality reduction. They tested on 10 different classes and achieved an accuracy of 90\% using SVM and 96.67\% using ANN. In \cite{Flores2021} the authors focused on the classification of four hand movements which we also presented in this paper (i.e., flexion, extension, opening, and closure). They tested with the wavelet transform, PCA, and univariate selection using the Chi-square function for feature extraction. They implemented SVM and ANN classifiers for the EMG signals. The best total classification accuracy they reported was 91\% using SVM and 89\% using ANN. The authors in \cite{Negi2016} used PCA and Uncorrelated Linear Discriminant Analysis (ULDA) to classify EMG signals relating to different hand movements. They used time-domain features in their approach and reached about 95\% classification accuracy for the PCA classification and about 96\% classification accuracy for the ULDA classification. In \cite{Leon2011}, they experimented with the percentages of training and testing data to observe the change in performance. The classification accuracy generally increased when the amount of testing data was reduced compared to the amount of training data. This approach affects the integrity of the overall classification accuracy results, however. In \cite{Yoo2019}, they focused on reducing the number of electrodes by using an optimal algorithm. The algorithm the authors used was Discriminative Feature-Oriented Dictionary Learning (DFDL). The classification accuracy using 6 channels was 94.1\%. Using 3 channels (as the case in our paper) the accuracy went down to 85.25\%. In \cite{Baygin2022}, they proposed a method called Frustum154 which utilizes Tunable Q-Factor Wavelet Transform (TQWT) as a decomposition method and Iterative Neighborhood Component Analysis (INCA) to choose the most discriminative features. The average classification accuracy reached over three datasets for six gestures was 96.38\%.

It is clearly visible that the methods provided in this paper yield similar or better results with fewer resources than what is current in the field. The total cost of the components of this prosthetic hand was 259.86 USD. Table \ref{table:cost} the cost breakdown. Moreover, we have a fully developed myoelectric prosthetic hand that is ready for testing on amputees.

\begin{table}[!t]
\caption{Cost breakdown for the designed system}
\centering
\begin{tabular}{|p{0.3\linewidth}||p{0.1\linewidth}|p{0.2\linewidth}|p{0.19\linewidth}|}
 \hline
 Unit & Quantity & Cost Per One Unit (USD) & Total Cost (USD) \\
 \hline\hline
 Arduino Mega 2560 & 1 & 38.50 & 38.50 \\ 
 \hline
 AD620 & 3 & 12.60 & 37.80 \\
 \hline
 AD8618 & 3 & 6.12 & 18.36 \\
 \hline
 \raggedright GWS MICRO 2BBMG & 3 & 15.85 & 47.55 \\
 \hline
 \raggedright 3D Printed Parts (PLA) & 1 & 80 & 80 \\
 \hline
 Steel Ball Bearing & 1 & 2.00 & 2.00 \\ 
 \hline
 FSR 400 (Force Sensors) & 5 & 6.95 & 34.75 \\ 
 \hline
 Coin Vibration Motor & 3 & 0.30 & 0.90 \\ 
 \hline
\end{tabular}
\label{table:cost}
\end{table}

\section{Conclusion}

The conductive fabric and carbon-silicon with gold coated copper wire electrodes gave constant performance and were easy and comfortable to wear. Using three input channels gave the best results while keeping the cost and complexity at minimum. Furthermore, connecting the amplifier directly to the electrodes limits the motion artifacts noise. After experimenting with different sampling rates and moving window size, 2000 Hz and 128 samples/window gave the best results while meeting hardware limitations. Training the neural network is considered of high computational complexity. Hence, to keep the systems’ cost as low as possible, the training was performed on an external software (MATLAB) leaving the classification to be performed on the Arduino. The time domain analysis yielded a better overall accuracy and a faster response and was consequently used as the median algorithm for the prosthetic. Tactile sensory feedback is a significant improvement to the prosthetic hand as a whole. The results provided in this paper regarding both the accuracy and complexity of the developed prosthetic hand provide a proof of concept for a low-cost myoelectric prosthetic hand.

\section{Acknowledgments}
The code developed and designs are provided in a repository at \url{https://github.com/M-DIAB/Myoelectric_Prosthetic_Hand}.

\bibliography{root}

\end{document}